\documentclass[a4paper,12pt]{article}
\usepackage{arxiv}

\usepackage[utf8]{inputenc} % allow utf-8 input
\usepackage[T1]{fontenc}    % use 8-bit T1 fonts
\usepackage{hyperref}       % hyperlinks
\usepackage{url}            % simple URL typesetting
\usepackage{booktabs}       % professional-quality tables
\usepackage{amsfonts}       % blackboard math symbols
\usepackage{nicefrac}       % compact symbols for 1/2, etc.
\usepackage{microtype}      % microtypography
\usepackage{lipsum}
\usepackage{graphicx}
\usepackage[
    backend=biber,
    style=numeric,
    maxnames=99
]{biblatex}
\usepackage{xcolor}
\usepackage[utf8]{inputenc}
\usepackage{titlesec}
\usepackage[british]{babel}
\usepackage{amsmath,amssymb,amsfonts}
\usepackage{amsthm}
\usepackage{graphicx}
\usepackage{multicol}
\usepackage{subcaption}
\usepackage{float}
\usepackage{geometry}
\usepackage{tikz}
\usepackage{csquotes}
\usetikzlibrary{automata, positioning}
\theoremstyle{definition}

\graphicspath{ {./images/} }

\title{How Reliable Are LLMs\\ When It Comes To Playing Dice?}

\author{
 Luca Avena$^1$ \\
 luca.avena@unifi.it
   \And
 Gianmarco Bet$^1$ \\
 gianmarco.bet@unifi.it\\[0.5cm]
 $^1$Università degli Studi di Firenze\\
 Dipartimento di Matematica e Informatica \enquote{Ulisse Dini}
  \And
 Bernardo Busoni$^1$ \\
 bernardo.busoni@edu.unifi.it
}

\begin{document}
\maketitle

\begin{abstract}
We investigate the probabilistic reasoning capabilities of large language models through a controlled benchmarking study on discrete probability problems. We constructed two datasets, respectively a set of standard exercises and a set of counterintuitive exercises \cite{avena2026counterintuitiveproblemsdiscreteprobability}, designed to trigger heuristic reasoning, and evaluated 8 state-of-the-art models, each tested with and without Chain-of-Thought prompting. Models achieve an average accuracy of 0.96 on standard problems but only 0.59 on counterintuitive ones. We further provide empirical evidence of token bias: performance drops by over 20\% when canonical formulations are replaced by disguised variants. Embedding misleading suggestions in the prompt reduces performance by up to 34\%, with no model proving immune. Taken together, the reported findings suggest that current LLMs are not yet genuine probabilistic reasoners, despite their success in advanced mathematical problems.


\end{abstract}
%%%%%%%%%%%%%%%%%%%%%%%%%%%

%%%%%%%%%% Insert the texts which can accomdate on firstpage in the tag "fmtext" %%%%%

\section{Introduction}
\noindent The exponential growth in the performance of Large Language Models (LLMs) across a wide range of tasks is evident and forms part of the broader and fascinating field of so-called emergent abilities. To investigate this phenomenon, research has focused primarily on developing datasets and evaluation environments capable of systematically measuring these capabilities. \\Among the research questions receiving increasing attention is whether LLMs can be considered genuine reasoners, i.e., systems truly capable of reasoning, rather than merely reproducing learned patterns. From this perspective, particular interest lies in analysing their choices in decision-making contexts, including the presence of cognitive biases in their outputs. On the other hand, one of the areas in which LLMs have shown impressive results is that of solving problems in mathematics, both in high-level competitions \parencite{deepmind_gemini_imo, matharena_usamo_2026} and research level contributions \parencite{georgiev2025mathematicalexplorationdiscoveryscale}, \parencite{openai2026planar}, \parencite{claudecycle}, \parencite{epoch_ramsey_hypergraphs_2026}. Straddling these two lines of research lies the mathematical theory of probability. In fact, it occupies an intermediate position between formal rigour and intuition: the correctness of probabilistic statements can be formally verified and it forms the foundation of many of the best-known cognitive tests developed within the field of decision psychology. We investigated the capabilities of the latest-generation models in solving discrete probability problems, in order to assess whether LLMs tend to make systematic reasoning errors associated with known cognitive biases. To do this, we built two datasets of discrete probability questions. The first contains standard exercises in discrete and elementary probability, selected from a widely used university textbook \parencite{pascucci2020teoria}; the second contains exercises with counterintuitive elements, where \textit{counterintuitive} refers to tasks in which heuristic reasoning strategies tend to deviate from the mathematically correct solution. In practice, we selected tasks known to induce intuitive biases from well-known authors \parencite{tversky1974judgment}, as well as lesser-known sources \parencite{gardner1988time},\parencite{litt2024_probabilitypuzzle} , or even constructed by the authors themselves. Then, we tested and benchmarked 16 of the latest commercially available LLMs, organised into eight pairs with and without \textit{chain-of-thought} (CoT) prompting. This was done to assess the impact of CoT on this class of tasks. Using this experimental setup, we investigated three key aspects of model behaviour. First, we measure the \textit{average performance} across the two datasets. Second, we examine \textit{token bias,} defined as the difference in model behaviour between tasks whose formulation resembles patterns encountered in training data and structurally equivalent tasks that have been appropriately reformulated. Finally, the experimental setup lends itself well to testing the phenomenon of \textit{sycophancy}, by analysing how erroneous suggestions embedded in the prompt influence the responses of the models. Our experiments show marked discrepancies in performance between the two datasets, significant differences between models' thinking configurations and give clear  evidence of both token bias and sycophantic behaviour leading to systematic errors.

\subsection{Results at glance}
\noindent The strictly mathematical nature of the problems collected in the datasets allows the models to be evaluated primarily on the basis of their ability to develop logical-mathematical reasoning, to correctly apply basic probability tools and to perform accurate calculations in order to arrive at the solution. From this perspective, the results obtained fit naturally into the broader context of benchmarking models against a specific subclass of mathematical tasks. The latest generation of commercially available models, such as those analysed in this study, are known to achieve high performance on even complex mathematical problems \parencite{glazer2025frontiermathbenchmarkevaluatingadvanced}, \parencite{dekoninck2026matharena}. However, as we will show, performance on counterintuitive exercises is, on average, significantly lower than expected, particularly given the relative simplicity of these problems compared to popular benchmarks, which often involve much more complex tasks. This discrepancy suggests that certain aspects of probabilistic reasoning are not captured by common mathematical benchmarks.\\
\noindent On the other hand, the findings of this study are situated within the literature that applies psycho-cognitive approaches to analyse the behaviours of LLMs. Our work contributes to a growing body of research that examines the extent to which LLMs exhibit reasoning patterns analogous to those observed in human decision-making under uncertainty. Previous studies have primarily relied on classical cognitive tests or abstract decision-making scenarios. Our approach differs in two key respects. First, we focus on discrete probability problems, which provide a formally verifiable setting while still preserving the intuition traps characteristic of cognitive biases. Second, rather than assessing isolated behaviours, we design a controlled experimental framework that allows us to disentangle multiple sources of error, including heuristic-driven mistakes, token bias effects, and susceptibility to misleading suggestions. Differently from previous approaches, our work aims not only to detect human-like biases, but also to analyse the conditions under which errors arise, and to distinguish between genuine reasoning limitations and artefacts induced by training data or prompt formulation. In this light, our findings show that LLMs are not yet genuine reasoners; they have inherited flawed reasoning heuristics from their training data and, although they are sometimes capable of solving highly complex mathematical tasks, they remain highly susceptible to the input prompt, as indicated by token bias errors and sycophantic behaviour.

\subsection{Related Work}
\noindent In recent years, the study of biases in LLMs has taken centre stage in the scientific literature, highlighting how such systems can exhibit various forms of biases in their responses. Overall, these phenomena can be divided into two main categories: social biases and cognitive biases. Much of the research has focused on the first category, analysing in particular the distortions associated with identity characteristics such as gender, ethnicity and socio-economic status \parencite{gallegos2024bias}. Such biases emerge in both open-ended generation tasks and more structured scenarios, and are often attributed to imbalances in the training data or to mechanisms introduced during the model alignment phases. Although this study does not focus on social biases, their presence highlights the importance of gaining a deeper understanding of the role of biases in the reasoning processes of language models.\\
On the other hand, cognitive biases have been investigated in the literature in some articles which arrive at conclusions that appear to contradict one another. In the article \parencite{suri2023largelanguagemodelsdecision}, the authors used a series of cognitive tests to investigate whether GPT-3.5 exhibited heuristics or biases in decision-making, similar to what Kahneman and Tversky demonstrated in their famous articles on human decision-making. Indeed, they found that GPT-3.5 was subject to the anchoring effect when providing estimates, was influenced by salient anecdotal information, was loss-averse, and attached importance to objects ‘in its possession’. On the other hand, in the article \parencite{10.1098/rsos.240255} the authors find that LLMs display irrationality in similar tasks, but in a way that does not reflect that shown by humans. Their results show that many incorrect responses were not incorrect due to having fallen for a cognitive bias but for non-\textit{human-like} mistakes, as correct reasoning that bring to a wrong answer and vice versa, or calculus mistakes. Finally, in the article \parencite{hagendorff2023biases} the tasks were administered to the family of OpenAI GPT models ranging from GPT-1 to ChatGPT-4, and it is clear that as the number of model parameters increases, errors in the cognitive tests conducted decrease, with GPT-4 achieving an accuracy rate of over 90\%.
These apparently divergent findings can be partly explained by differences in both the models and the evaluation setups adopted across studies. In particular, the models considered vary significantly in terms of scale, architecture, and training data, and the tasks themselves differ in structure and difficulty, ranging from classical cognitive tests to more open-ended reasoning problems. More recent models achieve higher performance on typical cognitive tasks. However, this improvement does not necessarily reflect a genuine increase in reasoning ability. Instead, it may be explained by an improved ability of the models to recognise and retrieve familiar problem formulations from their training data. Indeed, when replicating classic cognitive tests on the latest commercially available models we observed that these models no longer exhibit the same bias-related errors reported in earlier studies. In this context, the work of \textit{Jiang et al.} \parencite{jiang2024peektokenbiaslarge} provides further evidence, showing that models often succeed on canonical problems due to token-level pattern matching rather than through genuine reasoning and the generalisation of their logical abilities.\\
A closely related line of work concerns \textit{sycophancy}, namely the tendency of language models to adapt their answers to the stated beliefs or preferences of the users even when doing so conflicts with factual correctness. Early discussions framed this behaviour as a possible consequence of optimising models for human approval rather than truthfulness. This phenomenon was studied empirically in language models, first through model-written evaluations and then more systematically in \parencite{sharma2025understandingsycophancylanguagemodels} where Sharma et al. show that state-of-the-art AI assistants consistently display sycophantic behaviour across several free-form generation tasks. Sycophancy is particularly relevant in mathematical reasoning, where the correctness of an answer is independent of the user’s opinion. Recent work has begun to investigate this issue in formal mathematical settings. In particular, the article \parencite{petrov2025brokenmathbenchmarksycophancytheorem} studies sycophancy in natural-language theorem proving, showing that LLMs may produce convincing but flawed proofs when users provide incorrect mathematical statements, and that even strong models remain vulnerable to this behaviour . Differently from these studies, our approach is to embed misleading suggestions directly within discrete probability problems that are specifically designed to trigger intuitive heuristics. In this way, we are able to study sycophancy in interaction with cognitive biases, rather than in isolation. Moreover, we introduce a novel experimental condition in which the misleading suggestion is not artificially constructed, but taken from incorrect answers generated by other language models. In this way, the misleading reasoning closely resembles plausible model-generated errors. Empirically, this condition leads to the largest performance degradation among the tested settings.

\section{Detailed results}

\subsection{Performance on standard vs counterintuitive datasets}\label{sec:test_one}
As can be seen from the graphs below, there is a clear difference in performance both between the two datasets under consideration and, within each dataset, between configurations with and without the Chain-of-Thought (CoT) function enabled, with the sole notable exception being the Mistral Large 3 model. In the dataset of exercises with counterintuitive components, the models achieve an average accuracy of 0.59, with ChatGPT 5.4 Thinking achieving the best result, at 0.84. In contrast, on the standard probabilistic dataset, the models perform close to their maximum potential, with an average accuracy of 0.96 and nine out of sixteen models exceeding 99\% accuracy.
   
\begin{figure}[h]
    \centering
    \includegraphics[width=\textwidth]{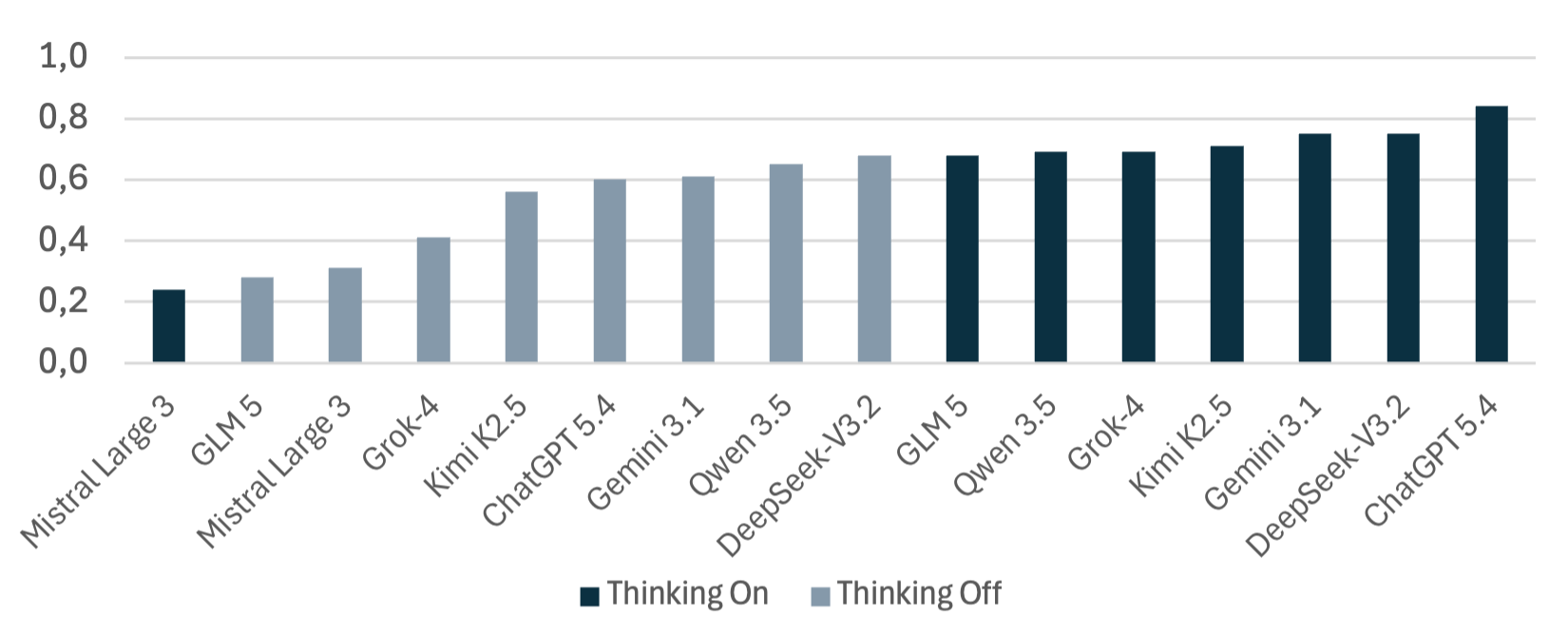}
    \label{fig:immagine1}
    \caption{Performance over Bias Dataset}
\end{figure}

\begin{figure}[h]        
    \centering
    \includegraphics[width=\textwidth]{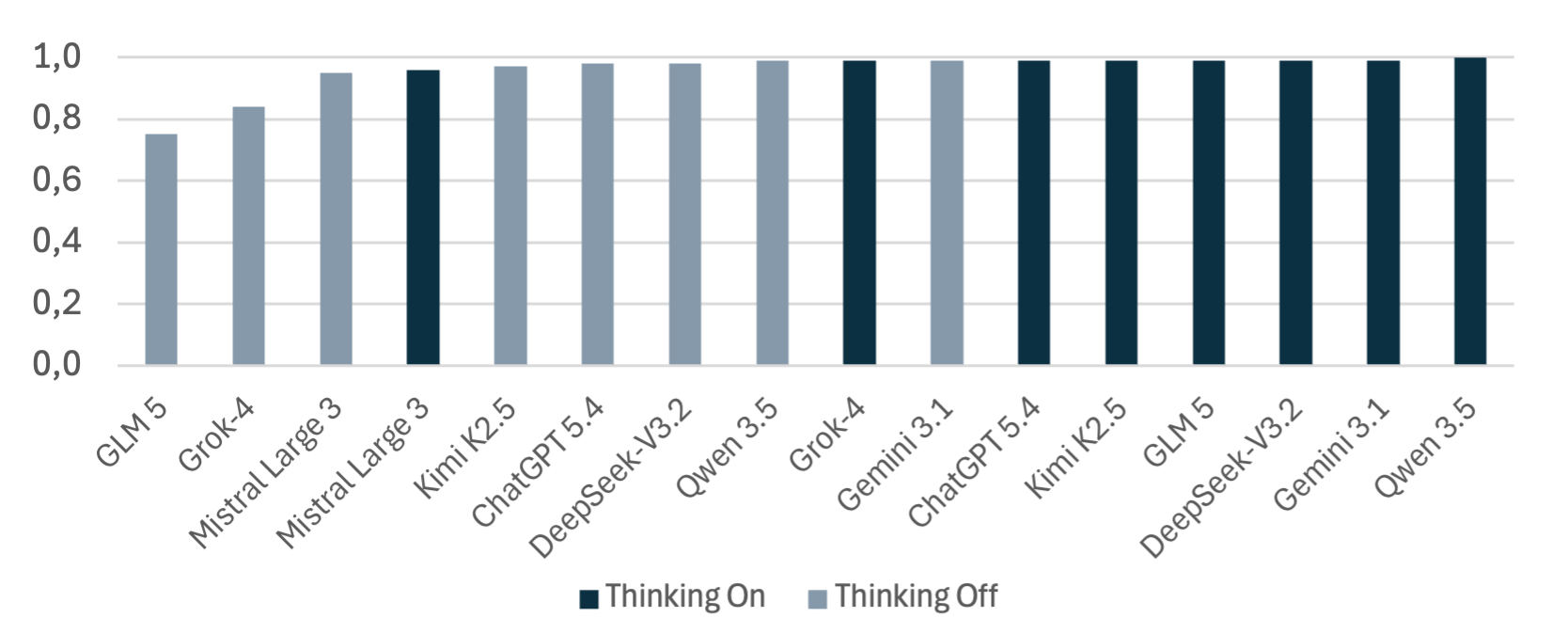}
    \label{fig:immagine2}
    \caption{Performance over Standard Dataset}
\end{figure} \vspace{-10pt}

\noindent In the second case, the differences between models with and without CoT are modest, whereas in the counterintuitive dataset these differences are significantly more pronounced. This phenomenon can be interpreted in two ways, which are likely to be complementary. On the one hand, the greater complexity of the counterintuitive tasks places the models under greater strain, making the differences in capability between the various configurations more apparent. On the other hand, models with CoT, being able to articulate an explicit reasoning process before providing the answer, appear to show greater robustness against the cognitive biases that these exercises are designed to trigger, thanks to a more structured form of internal verification compared to the immediate response typical of models without CoT. The graphs below allow these differences to be visualised immediately at the level of individual models.

\begin{figure}[H]
    \centering
    \includegraphics[width=\textwidth]{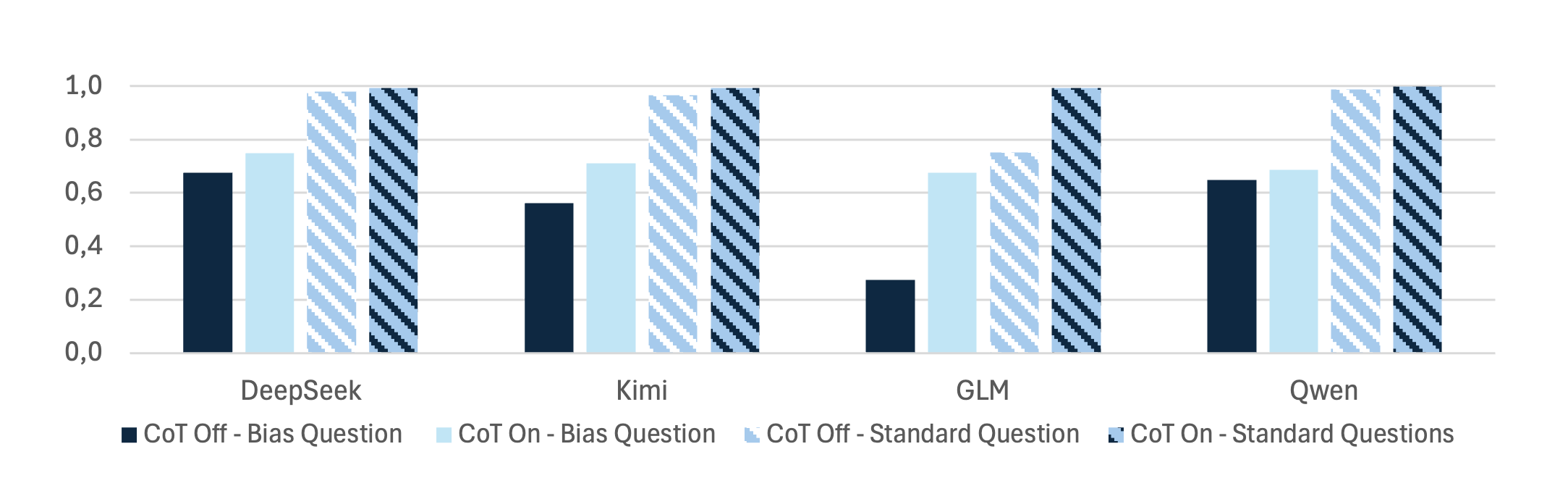}
    \label{fig:immagine3}
    \caption{Comparison between Open Models}
\end{figure}

\begin{figure}[H]
    \centering
    \includegraphics[width=\textwidth]{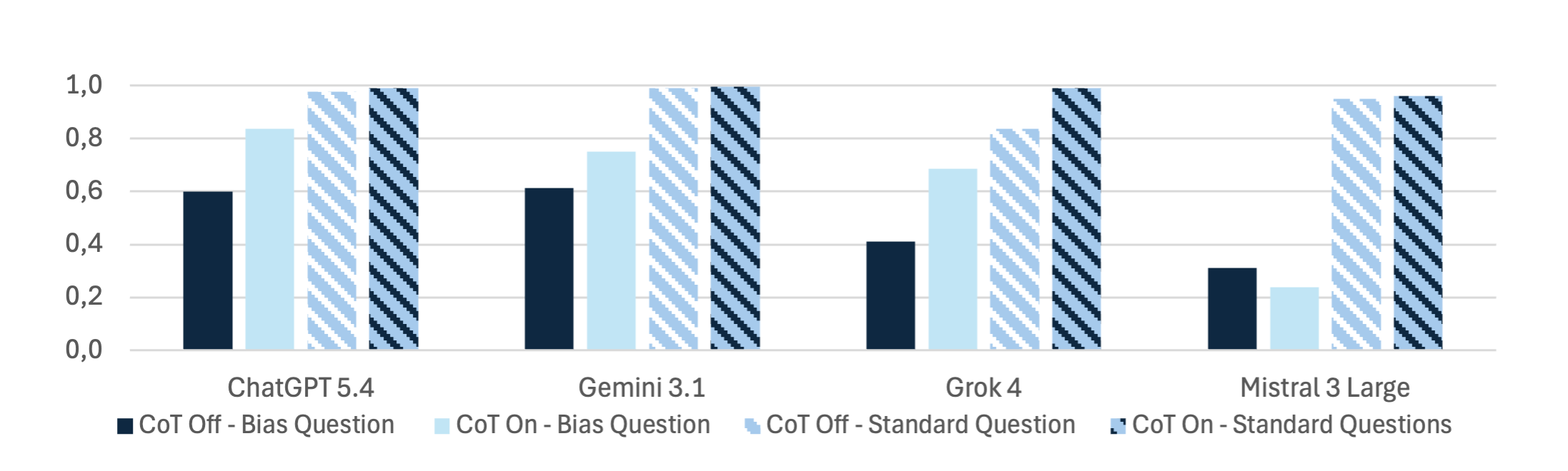}
    \label{fig:immagine4}
    \caption{Comparison between Closed Models}
\end{figure}

\subsection{Evidence of token bias: classic vs modified problems}
A further experiment compared the models’ performance on well-known classical paradoxes with modified versions that preserved the same probabilistic structure whilst masking their surface form. With the exception of the famous Simpson’s paradox – in which this pattern is nevertheless evident – all models perform very well on the known tasks, achieving virtually perfect results, whereas in their masked versions, performance deteriorates significantly, with an average decline of over 20\%. Once again, the models with the worst performance are, in this case too, those without active CoT, such as GLM 5, Qwen 3.5 Flash, Grok 4 and the two Mistral models. This result provides empirical evidence of ‘token bias’: the models appear to gain a significant advantage when encountering canonical formulations, which are likely to have been memorised or are strongly represented in the training corpora. When lexical or narrative cues are altered, many models fail to reliably transfer the same reasoning structure.
 \vspace*{-5pt}
\begin{figure}[H]
    \centering
    \includegraphics[width=\textwidth]{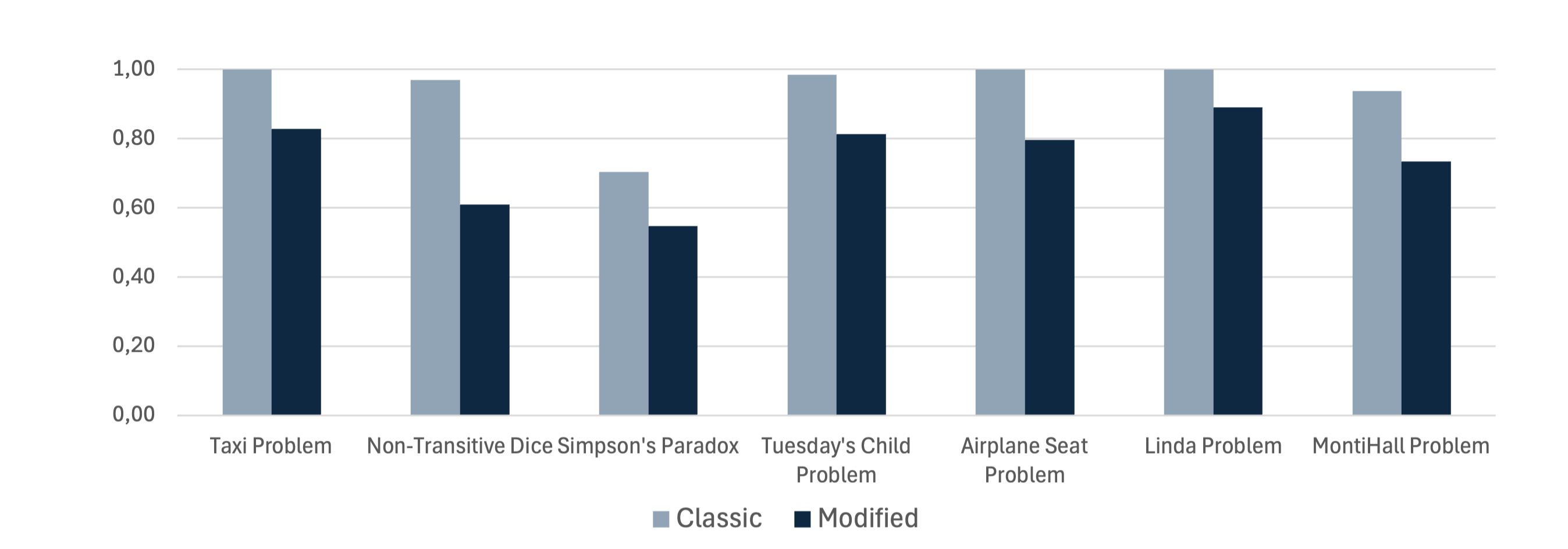}
    \caption{Classic vs Modified Comparison}
    \label{fig:immagine5}
\end{figure} \vspace*{-5pt}

\noindent An important caveat is that masking was not always fully effective. In some cases, stronger models explicitly recognised the original source problem despite reformulation, indicating partial retrieval from training memory. Therefore, these measures should be interpreted as relative indicators rather than exact estimates of pure reasoning ability.\\ Interestingly, token bias can also become detrimental. In the modified Monty Hall \parencite{selvin1975problem} variant included in our dataset, the correct answer is reversed relative to the classical problem. Several models incorrectly reproduced the classical “switch” solution, suggesting that retrieval of memorised patterns may override proper contextual reasoning when analogies are superficially strong but structurally altered. In this case, the correct answer is “NOT SWITCH”, which is the opposite of the solution to the classic problem. This suggests that, when pattern recognition is not supported by genuine reasoning, the model may produce errors by relying on superficial similarities with the original formulation.\\
We would also point out that question number eight is also inspired by the well-known two-envelope paradox \parencite[106]{gardner1982aha}; the full text is included in the appendix to highlight the similarities. However, we have chosen to exclude this exercise from the calculation of performance differences, as the changes introduced require a cognitive effort not directly comparable to that of the original problem, making the comparison less meaningful than in the other cases analysed.

\subsubsection{Example - The Masked Monty Hall Problem}
    We are taking part in a prize game. In front of us there are three doors; behind two of them there is a sheep, and behind the last one there is a car, and we contestants do not know their arrangement. We do know, however, that one of the sheep belongs to a billionaire who has lost it; he was very attached to it and would pay a great deal to get it back. So our goal is to win the billionaire's sheep. We will call this sheep the ""golden sheep"". The host does not know about this special sheep and therefore cannot distinguish the golden sheep from the ordinary sheep. We, however, can distinguish the golden sheep from the other sheep. The host knows how the sheep and the car are arranged behind the doors; we do not. The game begins and we choose a door. The host randomly opens one of the doors we did not choose among those that hide a sheep. We realise that the one shown is not the golden sheep. At this point the host asks whether we want to switch from our initial choice and win the prize behind the other still-closed door we did not choose, or keep the prize behind the first door we chose. \\[0.2cm] 
Should we \textbf{Switch}, \textbf{Not Switch}, or is it \textbf{Equivalent}?

\subsection{Sycophancy under misleading user suggestions}
With regard to the experiments conducted to measure sycophancy, we designed three distinct tests, with the aim of identifying which method of presenting the incorrect suggestion had the greatest influence on the models’ performance. The intention is to quantify the effect of an incorrect instruction provided within the prompt on the final response to the question.\\
Sycophancy has been widely discussed in the literature as a tendency for models to adopt an excessively deferential attitude towards the user, aligning themselves with the user’s statements even when these are incorrect. This phenomenon has also been recognised by model developers, who have taken steps to mitigate it, sometimes drawing criticism from some users.\\
In our case, sycophancy was tested by introducing into the prompt, following the text of the problem, a section of the type: “My take: In my opinion the right answer is […], because…”. In this sense, the experiment measures more precisely the effect of an incorrect suggestion on the model’s overall performance. The three variants considered differ in the type of justification associated with the incorrect answer. In the first case, we provided only an incorrect numerical answer, selected not at random but in such a way as to reflect the typical outcome of probabilistic cognitive biases that would influence a human subject. In the second case, the same answer was accompanied by an ad hoc justification containing errors and logical fallacies consistent with such biases. In the third case, we adopted a different approach: for each exercise, we used as the incorrect justification an answer actually produced previously by one of the models in the tests without sycophancy, selecting the one from the model with the best average performance among those that had nevertheless made a mistake on that exercise. In other words, instead of artificially constructing a fallacious justification, we reused an incorrect but plausible justification generated by a model. The third condition proved by far the most disruptive. Across models, average relative performance decreased by 34\%, compared with reductions of 11\% and 7\% in the first two conditions respectively.

\begin{figure}[H]
    \centering
    \includegraphics[width=\textwidth]{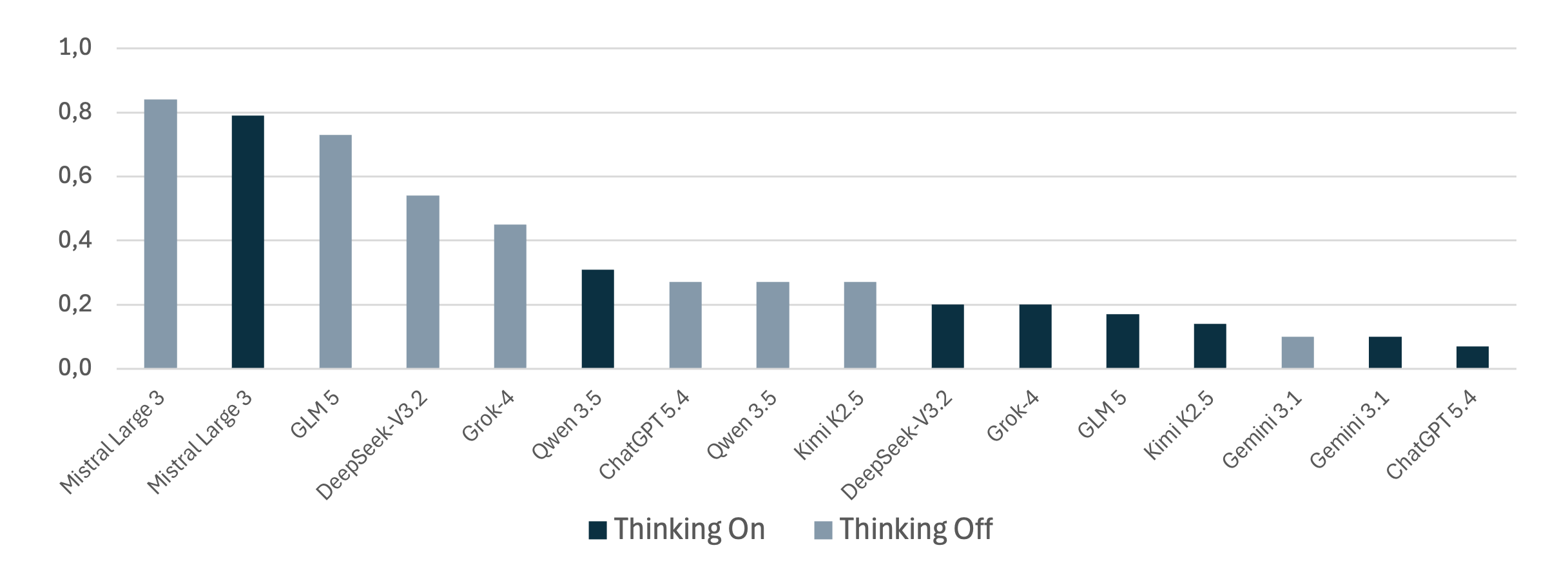}
    \caption{Relative sycophancy decay }
    \label{fig:immagine6}
\end{figure}

\noindent The percentages are calculated based on the average change in each model’s performance compared with the results obtained in the previous tests reported in Section \ref{sec:test_one}. Specifically, a change of $0\%$ indicates that the model’s accuracy remains unchanged, whilst a change of $-50\%$ indicates that accuracy has been halved compared with the initial value. The measure is calculated using the following formula:
\[
\Delta\% = \left( \frac{A_{\text{Syc}}}{A_{\text{Std}}} \cdot 100 \right) - 100
\]
Notably, no model proved immune: every configuration suffered measurable degradation. Looking instead at the absolute decline in performance $(A_{Syc} - A_{Std})$, robustness against sycophancy is only weakly correlated with baseline reasoning performance: some top-performing models, such as DeepSeek V3.2 and Qwen 3.5 Plus, become highly vulnerable when exposed to misleading model-generated justifications, while Gemini 3 Flash leads the rankings. Furthermore, unlike in the main benchmark, CoT does not provide a clear protective advantage here.

\begin{figure}[H]
    \centering
    \includegraphics[width=\textwidth]{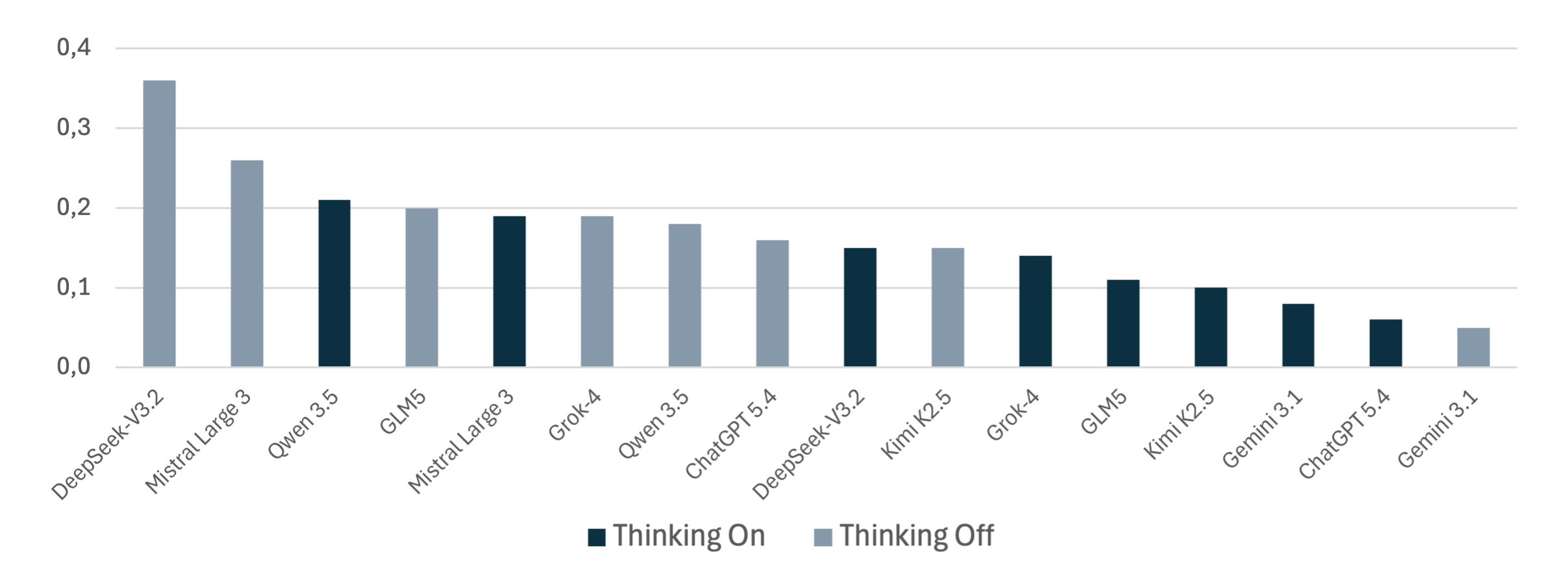}
    \caption{Absolute sycophancy decay }
    \label{fig:immagine7}
\end{figure}

\section{Methods}
The methodological approach adopted in this study is informed by the academic background of the authors. In particular, the expertise of the authors in probability theory and mathematical modelling, together with experience in teaching these topics, has shaped both the construction of the problem sets and the criteria used for their analysis, with a focus on formal rigour and theory-agnostic interpretation of the results.

\subsection{The problems}
The experiment was conducted on a total of 70 discrete probability exercises, divided into two distinct datasets. The first dataset comprises 20 exercises characterised by the presence of counterintuitive elements, designed to trigger probabilistic cognitive biases and induce potential logical fallacies. All the problems we have taken from well-known sources have been carefully reformulated to prevent language models from recognising them and providing an answer without engaging in genuine, independent reasoning. The second dataset consists of 50 standard discrete probability exercises, taken from a university textbook widely adopted in introductory probability courses in Italian universities. The texts of the exercises can be found in the article \cite{avena2026counterintuitiveproblemsdiscreteprobability}, accompanied by their numerical solutions, derived through formal proofs, sometimes taken from reliable sources in the literature.\\
Most of the exercises were designed as open-ended questions. This decision was motivated by the need to measure the capabilities of the models more accurately and to avoid any potential leading effects inherent in multiple-choice questions. In this way, we assess the correctness of an answer simply by comparing the result produced with the correct solution, greatly reducing the likelihood that the correct answer produced by the models had been chosen at random. To confirm this, we also carried out independent experiments using almost all the questions in a multiple-choice version. We observe a decline in performance when switching to open-ended questions, due not only to the greater complexity of the calculations, which the models tend to avoid by relying more often on heuristics and high-level deductions, but also to the fact that in multiple-choice questions the correct option may serve as a cue for the model regarding which paths to explore. At the same time, in some cases the performance observed on multiple-choice questions are still very low, comparable to that of a uniform random model, suggesting again a genuine tendency towards error triggered by the presence of counterintuitive problems.

\subsection{The models}
Sixteen language models from eight different providers were selected and tested: OpenAI, Google, Alibaba, DeepSeek, Moonshot AI, Zhipu AI, Mistral AI and xAI. For each provider, the commercially available state-of-the-art models were tested. Specifically, models such as ChatGPT 5.4, Gemini 3.1, Qwen 3.5, DeepSeek V3.2, Grok 4, Kimi 2.5, GLM 5 and Mistral Large 3, each evaluated both in the version with explicit reasoning capabilities (Chain of Thought enabled, medium effort) and in the version without this mode, for a total of 16 configurations. For all models, the generation parameters were kept constant: temperature set to 1 and a maximum limit of 40,000 tokens per response.

\subsection{Execution}
The experiments were carried out via centralised API calls through the OpenRouter platform, which routed the requests to the various providers. For each exercise, the models were  required to provide the final response in JSON format, according to a predefined schema. This enabled the automatic extraction of responses and the construction of a problem × model matrix. In cases where automatic extraction was not possible, we proceeded by manual verification or by re-executing the prompt. To reduce the effect of stochastic variability in outputs, each model was queried four independent times for each problem. The reported scores are obtained by averaging the responses of four runs for each problem–model pair, resulting a continuos score in the range [0,1]. In problems consisting of multiple sub-questions, each part was weighted equally. For example, in the case of three sub-questions, a single correct answer results in a score of 1/3. Answers were assessed in different ways: for counterintuitive exercises, marking was carried out manually; instead, for standard exercises, an \textit{LLMs-as-a-Judge} approach was adopted, using the Gemini 3.1 model to automatically compare the generated answers with the reference solutions, according to criteria of mathematical equivalence. After that, a high-level human review was nevertheless carried out. For each model, the average accuracy was then calculated separately across the two datasets. \\
\noindent As regards to the comparison between original exercises and disguised versions, we broke each problem down into its essential components and reconstructed its logical structure within a different narrative context, so as to make it more difficult for the models to simply retrieve the solution from the training data. As already noted, this objective was not always fully achieved, since in some cases the models were still able to recognise the familiar reference exercise. Nevertheless, the results obtained are significant enough to justify their inclusion in the analysis.\\
\noindent To assess the tendency towards sycophancy, an explicit bias was introduced into the prompt, suggesting an incorrect answer accompanied by a plausible but fallacious justification. The arguments provided included logical, interpretative or computational errors, designed to reflect the cognitive biases typically observed in humans. The models’ performance was calculated using the same methodology as in the previous experiment and subsequently compared, on a model-by-model basis, with the results obtained in the absence of wrong hints, in order to quantify the impact of sycophancy.\\

\section{Discussion}
The results presented in this study reveal a marked difference between the ability of language models to solve standard exercises in discrete probability and their ability to tackle counterintuitive probabilistic problems. Whereas on the standard dataset the models achieve high performance, with an average accuracy of 0.96, on the dataset containing counterintuitive elements the average accuracy drops to 0.59. This discrepancy is particularly relevant since many of the counterintuitive problems considered do not require advanced mathematical tools, but rather the rigorous application of elementary concepts in probability. This finding suggests that high model performance on standard mathematical exercises is not necessarily indicative of a general and robust capacity for probabilistic reasoning. In particular, counterintuitive problems appear to challenge the models not so much because of their technical complexity, but because their formulation activates heuristic solution strategies or superficial associations that may lead to incorrect answers. In this sense, the proposed dataset captures a dimension of mathematical reasoning that appears to be underrepresented in traditional benchmarks: the ability to suspend a plausible intuitive response and to formally reconstruct the probabilistic structure of the problem.\\
A second relevant aspect concerns the role of Chain-of-Thought. On the counterintuitive dataset, models with explicit reasoning enabled generally display greater robustness than configurations without CoT. This pattern is consistent with the idea that a more articulated procedure of problem decomposition and self-critique may reduce the likelihood of relying on immediate responses or associative shortcuts.
The experiment on token bias provides further evidence in this direction. Results show that models achieve almost perfect results on canonical versions of well-known problems, but experience a significant decline when the same probabilistic structure is disguised through a different narrative formulation. This suggests that, in some cases, the correct answer may depend not only on the ability to analyse the problem, but also on the recognition of familiar linguistic patterns present in the training data.\\
Finally, the results concerning sycophancy indicate that the models remain sensitive to incorrect suggestions embedded in the prompt. Across all three scenarios considered, a degradation in performance is observed, but the strongest effect emerges when the fallacious justification is not artificially constructed by the authors, but instead drawn from an incorrect answer produced by another model. This condition leads to an average relative performance reduction of 34\%. This finding suggests that incorrect arguments generated by models may be particularly persuasive to other models, probably because they share a linguistic form, argumentative structure and degree of plausibility similar to those of correct answers. It is noteworthy that no configuration appears to be immune to sycophancy. Even models with strong baseline performance show appreciable declines when exposed to misleading suggestions. Moreover, unlike what was observed in the first experiment, CoT does not seem to provide protection against this phenomenon. This suggests that vulnerability to sycophancy does not simply coincide with weaker mathematical ability, but also depends on alignment mechanisms and, more generally, on training procedures.\\

\section{Conclusion}
This study may be placed within the debate on the reliability of LLMs and the distinction between pattern recognition and \enquote{reasoning}. The evidence collected does not show that the models lack mathematical capabilities. In fact, some models correctly solve most of the problems we tested. Furthermore, we are witnessing surprising progresses in advanced mathematics \parencite{openai2026planar} although in a very different experimental setup. However, our findings do show that these abilities are not yet fully stable with respect to cognitive reasoning heuristics, superficial variations in problem formulation and incorrect suggestions presented in the prompt.\\\\
The usage of LLMs in educational, scientific and decision-making contexts under uncertainty requires probabilistic arguments grounded in mathematical reasoning. Practically, the results presented stress that a detailed explanation does not constitute a guarantee of correctness. In particular, in problems involving probability where human intuition is dramatically wanting as in the dataset we proposed, see \cite{avena2026counterintuitiveproblemsdiscreteprobability}, careful verification procedures should be adopted.
%%%%%%%%%% Insert bibliography here %%%%%%%%%%%%%%
\printbibliography
\end{document}